\documentclass[conference]{IEEEtran}
\IEEEoverridecommandlockouts
\usepackage{cite}
\usepackage{amsmath,amssymb,amsfonts}
\usepackage{algorithmic}
\usepackage{graphicx}
\usepackage{textcomp}
\usepackage{xcolor}

\usepackage{grffile} %
\usepackage[T1]{fontenc}
\usepackage[utf8]{inputenc}

\usepackage{bm} %
\usepackage{subcaption} %
\usepackage[hidelinks]{hyperref} %
\usepackage{braket} %
\usepackage{multirow} %
\usepackage{tikz}
\usepackage{tikz-cd}
\usetikzlibrary{shapes, arrows, positioning, decorations.pathreplacing, calc, external}
\tikzexternalenable
\tikzset{
  shift left/.style ={commutative diagrams/shift left={#1}},
  shift right/.style={commutative diagrams/shift right={#1}},
  -|/.style={to path={-| (\tikztotarget)}},
  |-/.style={to path={|- (\tikztotarget)}}
}

\def\BibTeX{{\rm B\kern-.05em{\sc i\kern-.025em b}\kern-.08em
    T\kern-.1667em\lower.7ex\hbox{E}\kern-.125emX}}

\newcommand{\x}{x}
\newcommand{\y}{y}
\newcommand{\xtilde}{\tilde{x}}
\newcommand{\ytilde}{\tilde{y}}
\newcommand{\Q}{Q}
\newcommand{\z}{z}
\newcommand{\ztilde}{\tilde{z}}
\newcommand{\s}{\sigma}
\newcommand{\stilde}{\tilde{\s}}
\newcommand{\fa}{f_{a}}
\newcommand{\fs}{f_{s}}
\newcommand{\fha}{f_{ha}}
\newcommand{\fhs}{f_{hs}}
\newcommand{\vone}{\textit{shallow}}
\newcommand{\Vone}{\textit{Shallow}}
\newcommand{\vtwo}{\textit{deep}}
\newcommand{\Vtwo}{\textit{Deep}}
\newcommand{\N}{N}
\newcommand{\NI}{\N_{1}}
\newcommand{\NII}{\N_{2}}
\newcommand{\NIII}{\N_{3}}
\newcommand{\Wi}{W}
\newcommand{\He}{H}
\newcommand{\De}{D}
\DeclareMathOperator{\FL}{FL}
\newcommand{\xt}{\x_t}
\newcommand{\xtildet}{\xtilde_t}
\newcommand{\alphat}{\alpha_t}
\newcommand{\thr}{t}
\newcommand{\thrstar}{t^\star}
\newcommand{\dist}{d}

\newcommand{\Hea}{H}
\newcommand{\condbaseline}{\textbf{\hyperref[ssec:modeltypes]{c1}}}

\newcommand{\condhyper}{\textbf{\hyperref[ssec:modeltypes]{c2}}}
\newcommand{\condhypersafe}{\textbf{c2}}
\newcommand{\conddeep}{\textbf{\hyperref[ssec:transforms]{c3}}}
\newcommand{\conddeepsafe}{\textbf{c3}}
\newcommand{\condalpha}{\textbf{\hyperref[ssec:focalloss]{c4}}}
\newcommand{\condalphasafe}{\textbf{c4}}
\newcommand{\condoptimal}{\textbf{\hyperref[ssec:optimal]{c5}}}
\newcommand{\condoptimalsafe}{\textbf{c5}}
\newcommand{\condsequential}{\textbf{\hyperref[ssec:sequential]{c6}}}
\newcommand{\condsequentialsafe}{\textbf{c6}}
\DeclareMathOperator*{\argmin}{arg\,min}
\newcommand\dstA{0.6cm}
\newcommand\dstB{0.0cm}
\newcommand\dstC{0.2cm}

\begin{document}

\title{Improved Deep Point Cloud Geometry Compression%
\thanks{Funded by ANR ReVeRy national fund (REVERY ANR-17-CE23-0020).}}

\author{\IEEEauthorblockN{Maurice Quach\IEEEauthorrefmark{1},
Giuseppe Valenzise\IEEEauthorrefmark{1},
Frederic Dufaux\IEEEauthorrefmark{1}}
\IEEEauthorblockA{\IEEEauthorrefmark{1}Universit\'e Paris-Saclay, CNRS, CentraleSup\'elec, Laboratoire des signaux et syst\`emes\\
91190 Gif-sur-Yvette, France}}

\maketitle

\begin{abstract}
Point clouds have been recognized as a crucial data structure for 3D content and are essential in a number of applications such as virtual and mixed reality, autonomous driving, cultural heritage, etc.
In this paper, we propose a set of contributions to improve deep point cloud compression, i.e.:
using a scale hyperprior model for entropy coding; employing deeper transforms; a different balancing weight in the focal loss; optimal thresholding for decoding; and sequential model training.
In addition, we present an extensive ablation study on the impact of each of these factors, in order to provide a better understanding about why they improve RD performance. 
An optimal combination of the proposed improvements achieves BD-PSNR gains over G-PCC trisoup and octree of 5.50 (6.48) dB and 6.84 (5.95) dB, respectively, when using the point-to-point (point-to-plane) metric.
Code is available at \url{https://github.com/mauriceqch/pcc_geo_cnn_v2}.
\end{abstract}


\section{Introduction}

Due to recent advances in visual capture technology, point clouds have been recognized as a crucial data structure for 3D content.
In particular, point clouds are essential for numerous applications such as virtual and mixed reality, sensing for autonomous vehicle navigation, architecture and cultural heritage, etc.
Point clouds are sets of 3D points identified by their coordinates, which constitute the geometry of the point cloud.
In addition, each point can be associated with attributes like colors, normals and reflectance.
Point clouds can have a massive number of points, especially in high precision or large scale captures. This entails a huge storage and transmission cost. As a result, Point Cloud Compression (PCC) is fundamental in practice.

The Moving Picture Experts Group (MPEG) is planning to release two PCC standards \cite{schwarz_emerging_2018}: Geometry-based PCC (G-PCC) and Video-based PCC (V-PCC).
G-PCC approaches PCC from a 3D perspective and compresses point clouds in their native form using 3D data structures such as octrees.
On the other hand, V-PCC approaches PCC from a 2D perspective, projects 3D data onto a 2D plane and makes use of video compression technology.
In order to evaluate test models, common test conditions (CTCs) \cite{noauthor_common_2020} were designed.
In this context, the point-to-point (D1) and the point-to-plane \cite{tian_geometric_2017} quality metrics (D2) are used for quantitative evaluation. Recently, deep point cloud compression (DPCC) methods have been proposed and shown to provide significant coding gains compared to traditional methodologies~\cite{quach_learning_2019,guarda_deep_2019}.

In this paper, we focus on lossy compression of static point cloud geometry using deep convolutional networks.
Specifically, we propose a set of contributions to improve RD performance and accelerate model training.
We then present an ablation study identifying key performance factors for DPCC.
In particular, we start from a baseline DPCC model~\cite{quach_learning_2019} and we consider the following improvements:
\begin{itemize}
  \item \textit{Entropy modeling}: we consider an hyperprior model to improve entropy coding.
  \item \textit{Deeper transforms} that compensate downsampling with progressively higher numbers of filters.
  \item \textit{Changing the balancing weight in the focal loss}: similar to~\cite{quach_learning_2019}, we cast PCC decoding as an unbalanced classification problem by optimizing a focal loss~\cite{lin_focal_2017}. Hence, we study the RD performance impact of the focal loss $\alpha$ parameter.
  \item \textit{Optimal thresholding for decoding}: in order to classify voxels as occupied or not, we propose an optimal thresholding approach that minimizes a given distortion metric (instead of a fixed threshold as in~\cite{quach_learning_2019}).
  \item \textit{Sequential training}: in order to reduce the computational complexity of training a network for each RD tradeoff, we propose a sequential training procedure. That is, we train a network corresponding to a given RD point by fine tuning the network trained from the previous RD point. This makes training times up to $8$ times faster compared to training independently and improves RD performance.
  \item \textit{Ablation study} An extensive ablation study evaluating the impact of each factor mentioned above on RD performance. The evaluated conditions are detailed in Table \ref{tbl:expcond}.
  \item \textit{Octree partitioning} An efficient octree partitioning algorithm that is significantly faster compared to recursive octree partitioning.
\end{itemize}

\section{Related Work}

Our research is related most closely to three research areas: static point cloud geometry compression, deep image compression and deep point cloud compression.

Static point cloud geometry compression methods are usually based on the octree structure \cite{schnabel_octree-based_2006}.
Indeed, octrees provide an efficient way of partitioning the 3D space and representing point clouds.
In particular, they are especially suitable for lossless coding in combination with octree entropy models~\cite{garcia_intra-frame_2018}.
However, lossy compression using octrees alone has poor performance as pruning octree levels decreases the number of points exponentially resulting in significant distortion.
To alleviate this issue, many solutions have been proposed such as triangle \cite{dricot_adaptive_2019} surface models, planar \cite{dricot_hybrid_2019} surface models, graph-based enhancement layers \cite{de_oliveira_rente_graph-based_2019} and volumetric functions \cite{krivokuca_volumetric_2018}.
The core idea is that by encoding approximations along a coarse octree, we can alleviate the shortcomings of the octree structure.
Different from previous work in this area, we study learned approximation models based on deep neural networks.

Deep image compression considers the use of deep neural networks for image compression.
An end-to-end image compression solution with joint RD optimization along with a learned entropy model has been proposed in \cite{balle_end--end_2017}, which also replaces (non-differentiable) quantization with uniform noise at training time.
As a follow-up of that work, a scale hyperprior model has been proposed in \cite{balle_variational_2018}.
The scale hyperprior enables the modeling of spatial correlations in the latent space; for each element, it uses a Gaussian distribution whose standard deviation is predicted by a dedicated network.
We design models for PCC using these learning-based entropy modeling techniques.

DPCC is a recent research avenue exploring the use of deep neural networks for PCC.
For lossy geometry coding, voxel-based DPCC methods have been shown to outperform traditional methods significantly \cite{quach_learning_2019, wang_learned_2019, guarda_deep_2019}.
For lossless geometry coding, deep neural networks have been used to improve entropy modeling \cite{huang_octsqueeze_2020}.
Also, DPCC for attributes has been explored by interpreting point clouds as a 2D discrete manifold in 3D space \cite{quach_folding-based_nodate}.
Closely related to our study, the behavior and performance of DPCC methods has been investigated in \cite{guarda_deep_2019}.
However, this particular study investigates the characteristics and RD impact of the latent space.
In contrast, we seek to understand and identify key performance factors for rate-distortion (RD) performance on a larger scale.

\section{Proposed Improvements}\label{sec:proposed}

\begin{table}
  \centering
  \begin{tabular}{|l|l|l|r|l|l|}
    \hline
    Name              & Model       & Transforms  & $\alpha$  & Threshold & Training    \\ \hline
    \condbaseline{}   & Baseline    & \Vone{}     & $0.90$    & Fixed     & Independent \\ \hline
    \condhyper{}      & Hyperprior  & ---         & ---       & ---       & ---         \\ \hline
    \conddeep{}       & ---         & \Vtwo{}    & ---       & ---       & ---         \\ \hline
    \condalpha{}      & ---         & ---         & $0.75$    & ---       & ---         \\ \hline
    \condoptimal{}    & ---         & ---         & ---       & Optimal   & ---         \\ \hline
    \condsequential{} & ---         & ---         & ---       & ---       & Sequential  \\ \hline
  \end{tabular}
  \caption{Experimental conditions evaluated in this study.}
  \label{tbl:expcond}
\end{table}

\begin{figure}
\centering
\begin{subfigure}[b]{0.45\columnwidth}
	\centering
	\begin{tikzpicture}[
			mainnode/.style={draw, rectangle, minimum width=0.75cm, minimum height=0.5cm, align=center,
				font=\footnotesize},
            mininode/.style={rectangle, fill=white},
		]
		\node (in) {$\x$};
		\node (fa) [mainnode, below=\dstC of in.south, anchor=north] {$\fa$};
		\node (quant) [mainnode, below=\dstA of fa.south, anchor=north] {$\Q$};
		\node (fs) [mainnode, below=\dstA of quant.south, anchor=north] {$\fs$};
		\node (out) [below=\dstC of fs.south, anchor=north, minimum height=0.6cm] {$\xtilde$};

		\draw [->] (in) -- (fa);
		\draw [->] (fa) -- node (y) [mininode] {$\y$} (quant);
		\draw [->] (quant) -- node (yhat) [mininode] {$\ytilde$} (fs);
        \draw [->] (fs) -- (out);
        
        \node (ac) [mainnode, right=\dstA of fs.center, anchor=west] {$A\smash{C_{\ytilde}}$};

        \node (bstr) [minimum width=11*0.0625cm, minimum height=0.6cm, inner sep=0pt, below=\dstC of ac.south, anchor=north] {\small{0101}};
        \path[->, shift left=.5ex]
            ([xshift=-.375ex, yshift=-.125ex]yhat.east) edge[-|] ([xshift=.25ex, yshift=-.5ex]ac.north)
            ([xshift=-.375ex, yshift=.45ex]ac.north) edge[|-] ([yshift=-.125ex]yhat.east)
            (ac) edge (bstr)
            (bstr) edge (ac);

	\end{tikzpicture}
	\caption{Baseline model.}
	\label{fig:baselinemodel}
\end{subfigure}
\begin{subfigure}[b]{0.45\columnwidth}
	\centering
	\begin{tikzpicture}[
			mainnode/.style={draw, rectangle, minimum width=0.75cm, minimum height=0.5cm, align=center,
				font=\footnotesize},
            mininode/.style={rectangle, fill=white}
		]
		\node (in) {$\x$};
		\node (fa) [mainnode, below=\dstC of in.south, anchor=north] {$\fa$};

		\node (y) [mininode, right=\dstA of fa.east, anchor=center] {$\y$};
		\node (fha) [mainnode, right=\dstA * 2 of y.center, anchor=west] {$\fha$};
		\node (quant2) [mainnode, below=\dstA of fha.south, anchor=north] {$\Q$};
		\node (fhs) [mainnode, below=\dstA of quant2.south, anchor=north] {$\fhs$};

		\node (quant) [mainnode, left=\dstA * 2 of fhs.west, anchor=center] {$\Q$};
		\node (yhat) [mininode, below=\dstA of quant.south, anchor=north] {$\ytilde$};
		\node (fs) [mainnode, left=\dstA of yhat.center, anchor=east] {$\fs$};
		\node (out) [below=\dstC of fs.south, anchor=north, minimum height=0.6cm] {$\xtilde$};

		\draw [->] (in) -- (fa);
		\draw (fa) -- (y);
		\draw [->] (y) -- (fha);
		\draw [->] (y) -- (quant);
		\draw [->] (fha) -- node (z) [mininode] {$\z$} (quant2);
		\draw [->] (quant2) -- node (zhat) [mininode] {$\ztilde$} (fhs);
		\draw (quant) -- (yhat);
		\draw [->] (yhat) -- (fs);
        \draw [->] (fs) -- (out);
        
        \node (ac1) [mainnode, right=\dstA*2 of yhat.center, anchor=west] {$A\smash{C_{\ytilde}}$};

        \node (bstr) [minimum width=11*0.0625cm, minimum height=0.6cm, inner sep=0pt, below=\dstC of ac1.south, anchor=north] {\small{0101}};
        \path[->, shift left=.5ex]
            (yhat) edge (ac1)
            (ac1) edge (bstr)
            (bstr) edge (ac1)
            (ac1) edge (yhat);
		\draw [->] (fhs) -- node (s) [mininode] {$\stilde$} (ac1);

        \node (ac2) [mainnode, right=\dstA of ac1.center, anchor=west] {$AC_{\ztilde}$};

        \node (bstr) [minimum width=11*0.0625cm, minimum height=0.6cm, inner sep=0pt, below=\dstC of ac2.south, anchor=north] {\small{0101}};
        \path[->, shift left=.5ex]
            ([xshift=-.375ex]zhat.east) edge[-|] ([yshift=-.375ex]ac2.north)
            ([xshift=-.125ex, yshift=.25ex]ac2.north) edge[|-] ([xshift=.375ex, yshift=-.25ex]zhat.east)
            (ac2) edge (bstr)
            (bstr) edge (ac2);

	\end{tikzpicture}
	\caption{Hyperprior model.}
	\label{fig:hyperpriormodel}
\end{subfigure}
\caption{Entropy models considered in this work. The $f$ functions are learned transforms, $\Q$ refers to quantization and $AC$ refers to arithmetic coding with its associated density model.}
\label{fig:modeltypes}
\end{figure}
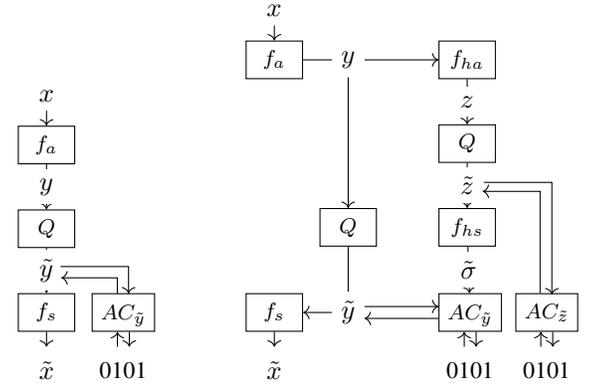

In this section we present different strategies to improve DPCC. We consider as baseline the network proposed in our preliminary work~\cite{quach_learning_2019} (denoted as \condbaseline{} in the following). In that work, we relied on \vone{} transforms to compress entire point clouds at once. However, this has a fundamental limitation in terms of memory usage, as it does not allow to compress large point clouds as those commonly used in MPEG CTCs.
Therefore, in this work we make use of octree partitioning to partition point clouds into blocks of size $64\times 64 \times 64$ voxels, which we have found to be a good trade-off between memory usage and coding performance. In the rest of the paper, we denote the different considered improvements  with \textbf{c2},$\ldots$, \textbf{c6}, which are summarized in Table~\ref{tbl:expcond}.

\subsection{Entropy modeling (\condhypersafe{})} \label{ssec:modeltypes}
We consider models that take the voxelized point clouds $\x$ and $\xtilde$ as input and output.
In particular, we consider a baseline model (Fig. \ref{fig:baselinemodel}) and an hyperprior model (Fig. \ref{fig:hyperpriormodel}).

The baseline model is based on an autoencoder architecture with an analysis $\fa$ and a synthesis transform $\fs$ \cite{balle_end--end_2017}.
$\y$ is modeled using a learned entropy model for each feature map.
The baseline model is expressed as follows
\begin{align}
\y &= \fa(\x) & \ytilde &= Q(\y) & \xtilde &= \fs(\ytilde).
\end{align}

We consider a scale hyperprior model \cite{balle_variational_2018} as a better entropy model for $\ytilde$.
Specifically, we model $\y$ with a zero-mean gaussian density model $\mathcal N(0, \stilde^2)$ where standard deviations $\stilde^2$ are predicted from $\y$ with $\stilde = \fhs(Q(\fha(\y)))$.
As a result, the spatial dependencies can be modeled better compared to the learned entropy model.
The hyperprior model is expressed as follows
\begin{align}
    \y &= \fa(\x) & \ytilde &= Q(\y) & \xtilde &= \fs(\ytilde) \\
    \z &= \fha(\y) & \ztilde &= Q(z) & \stilde &= \fhs(\ztilde)
\end{align}
where $\z$ is modeled with a learned density model for each feature map.

The compression model is trained using joint RD optimization with the loss function $R + \lambda D$.
For each RD tradeoff, we train a model with the corresponding $\lambda$ value resulting in transforms and entropy models specialized for this particular tradeoff.
The entropy $R$ is computed on $\ytilde$, and $\ztilde$ for the hyperprior model, using their associated entropy models.
Since the quantization operation $\Q$ is not differentiable, we use additive uniform noise during training in place of quantization as originally proposed in~\cite{balle_end--end_2017}.

\subsection{Deeper transforms (\conddeepsafe{})} \label{ssec:transforms}

\begin{figure}
\centering
\begin{subfigure}{0.32\columnwidth}
	\centering
	\begin{tikzpicture}[
			mainnode/.style={draw, rectangle, minimum width=1.5cm, minimum height=0.45cm, align=center,
				font=\tiny}
		]
		\node (in) {};
		\node (box1) [mainnode, below=\dstC of in.south, anchor=north] {C, $\N$, $9$, $\downarrow 2$};
		\node (box2) [mainnode, below=\dstB of box1.south, anchor=north] {C, $\N$, $5$, $\downarrow 2$};
		\node (box3) [mainnode, below=\dstB of box2.south, anchor=north] {C, $\N$, $5$, $\downarrow 2$};
		\node (out) [below=\dstC of box3.south, anchor=north] {};
		\node (stretch) [below=2.25cm of in.south, anchor=north] {};

		\draw [->] (in) -- (box1);
		\draw [->] (box3) -- (out);
	\end{tikzpicture}
	\caption{\Vone{} Analysis}
	\label{fig:transforms-analysis-v1}
\end{subfigure}
\begin{subfigure}{0.32\columnwidth}
	\centering
	\begin{tikzpicture}[
			mainnode/.style={draw, rectangle, minimum width=1.5cm, minimum height=0.45cm, align=center,
				font=\tiny}
		]
		\node (in) {};
		\node (box1) [mainnode, below=\dstC of in.south, anchor=north] {AB, $\NI$, $\downarrow 2$};
		\node (box2) [mainnode, below=\dstB of box1.south, anchor=north] {AB, $\NII$, $\downarrow 2$};
		\node (box3) [mainnode, below=\dstB of box2.south, anchor=north] {AB, $\NIII$, $\downarrow 2$};
		\node (box4) [mainnode, below=\dstB of box3.south, anchor=north] {C, $\NIII$, $3$};
		\node (out) [below=\dstC of box4.south, anchor=north] {};
		\node (stretch) [below=2.25cm of in.south, anchor=north] {};

		\draw [->] (in) -- (box1);
		\draw [->] (box4) -- (out);
	\end{tikzpicture}
	\caption{\Vtwo{} Analysis}
	\label{fig:transforms-analysis-v2}
\end{subfigure}
\begin{subfigure}{0.32\columnwidth}
	\centering
	\begin{tikzpicture}[
			mainnode/.style={draw, rectangle, minimum width=1.5cm, minimum height=0.45cm, align=center,
				font=\tiny}
		]
		\node (in) {};
		\node (box1) [mainnode, below=\dstC of in.south, anchor=north] {C, $\N$, $3$};
		\node (box2) [mainnode, below=\dstB of box1.south, anchor=north] {C, $\N$, $3$, $\downarrow 2$};
		\node (box3) [mainnode, below=\dstB of box2.south, anchor=north] {C, $\N$, $3$};
		\node (out) [below=\dstC of box3.south, anchor=north] {};
		\node (stretch) [below=2.25cm of in.south, anchor=north] {};

		\draw [->] (in) -- (box1);
		\draw [->] (box3) -- (out);
	\end{tikzpicture}
	\caption{Hyper analysis}
	\label{fig:transforms-hyper-analysis}
\end{subfigure}
\begin{subfigure}{0.32\columnwidth}
	\centering
	\begin{tikzpicture}[
			mainnode/.style={draw, rectangle, minimum width=1.5cm, minimum height=0.45cm, align=center,
				font=\tiny}
		]
		\node (in) {};
		\node (box1) [mainnode, below=\dstC of in.south, anchor=north] {C\textsuperscript{T}, $\N$, $5$, $\uparrow 2$};
		\node (box2) [mainnode, below=\dstB of box1.south, anchor=north] {C\textsuperscript{T}, $\N$, $5$, $\uparrow 2$};
		\node (box3) [mainnode, below=\dstB of box2.south, anchor=north] {C\textsuperscript{T}, $\N$, $9$, $\uparrow 2$};
		\node (out) [below=\dstC of box3.south, anchor=north] {};
		\node (stretch) [below=2.25cm of in.south, anchor=north] {};

		\draw [->] (in) -- (box1);
		\draw [->] (box3) -- (out);
	\end{tikzpicture}
	\caption{\Vone{} Synthesis}
	\label{fig:transforms-synthesis-v1}
\end{subfigure}
\begin{subfigure}{0.32\columnwidth}
	\centering
	\begin{tikzpicture}[
			mainnode/.style={draw, rectangle, minimum width=1.5cm, minimum height=0.45cm, align=center,
				font=\tiny}
		]
		\node (in) {};
		\node (box1) [mainnode, below=\dstC of in.south, anchor=north] {SB, $\NIII$, $\uparrow 2$};
		\node (box2) [mainnode, below=\dstB of box1.south, anchor=north] {SB, $\NII$, $\uparrow 2$};
		\node (box3) [mainnode, below=\dstB of box2.south, anchor=north] {SB, $\NI$, $\uparrow 2$};
		\node (box4) [mainnode, below=\dstB of box3.south, anchor=north] {C, $1$, $3$};
		\node (out) [below=\dstC of box4.south, anchor=north] {};
		\node (stretch) [below=2.25cm of in.south, anchor=north] {};

		\draw [->] (in) -- (box1);
		\draw [->] (box4) -- (out);
	\end{tikzpicture}
	\caption{\Vtwo{} Synthesis}
	\label{fig:transforms-synthesis-v2}
\end{subfigure}
\begin{subfigure}{0.32\columnwidth}
	\centering
	\begin{tikzpicture}[
			mainnode/.style={draw, rectangle, minimum width=1.5cm, minimum height=0.45cm, align=center,
				font=\tiny}
		]
		\node (in) {};
		\node (box1) [mainnode, below=\dstC of in.south, anchor=north] {C\textsuperscript{T}, $\N$, $3$};
		\node (box2) [mainnode, below=\dstB of box1.south, anchor=north] {C\textsuperscript{T}, $\N$, $3$, $\uparrow 2$};
		\node (box3) [mainnode, below=\dstB of box2.south, anchor=north] {C\textsuperscript{T}, $\N$, $3$};
		\node (out) [below=\dstC of box3.south, anchor=north] {};
		\node (stretch) [below=2.25cm of in.south, anchor=north] {};

		\draw [->] (in) -- (box1);
		\draw [->] (box3) -- (out);
	\end{tikzpicture}
	\caption{Hyper synthesis}
	\label{fig:transforms-hyper-synthesis}
\end{subfigure}
\begin{subfigure}{0.4\columnwidth}
	\centering
	\begin{tikzpicture}[
			mainnode/.style={draw, rectangle, minimum width=1.5cm, minimum height=0.45cm, align=center, font=\tiny},
			circlenode/.style={draw, circle, inner sep=1pt, align=center, font=\tiny}
		]
		\node (in) {};
		\node (box1) [mainnode, below=\dstC of in.south, anchor=north] {C, $\N$, $3$, $\downarrow 2$};
		\node (resid) [coordinate, below=\dstC/2 of box1.south, anchor=center] {};
		\node (residl) [left=1.1cm of resid.center, anchor=center] {};
		\node (residr) [coordinate, right=1.1cm of resid.center, anchor=center] {};
		\node (stretchresidl) [coordinate, left=1.1cm of resid.center, anchor=center] {};
		\node (box2) [mainnode, below=\dstC of box1.south, anchor=north] {C, $\N$, $3$};
		\node (box3) [mainnode, below=\dstB of box2.south, anchor=north] {C, $\N$, $3$};
		\node (add) [circlenode, below=\dstC of box3.south, anchor=north] {$+$};
		\node (residbr) [coordinate, right=1.1cm of add.center, anchor=center] {};
		\node (out) [below=\dstC of add.south, anchor=north] {};
		\node (stretch) [below=2.25cm of in.south, anchor=north] {};

		\draw [->] (in) -- (box1);
		\draw [->] (box1) -- (box2);
		\draw (resid) -- (residr);
		\draw (residr) -- (residbr);
		\draw [->] (residbr) -- (add);
		\draw [->] (box3) -- (add);
		\draw [->] (add) -- (out);
	\end{tikzpicture}
	\caption{Analysis block (AB)}
	\label{fig:transforms-analysis-block}
\end{subfigure}
\begin{subfigure}{0.4\columnwidth}
	\centering
	\begin{tikzpicture}[
			mainnode/.style={draw, rectangle, minimum width=1.5cm, minimum height=0.45cm, align=center, font=\tiny},
			circlenode/.style={draw, circle, inner sep=1pt, align=center, font=\tiny}
		]
		\node (in) {};
		\node (box1) [mainnode, below=\dstC of in.south, anchor=north] {C\textsuperscript{T}, $\N$, $3$, $\uparrow 2$};
		\node (resid) [coordinate, below=\dstC/2 of box1.south, anchor=center] {};
		\node (residl) [left=1.1cm of resid.center, anchor=center] {};
		\node (residr) [coordinate, right=1.1cm of resid.center, anchor=center] {};
		\node (box2) [mainnode, below=\dstC of box1.south, anchor=north] {C\textsuperscript{T}, $\N$, $3$};
		\node (box3) [mainnode, below=\dstB of box2.south, anchor=north] {C\textsuperscript{T}, $\N$, $3$};
		\node (add) [circlenode, below=\dstC of box3.south, anchor=north] {$+$};
		\node (residbr) [coordinate, right=1.1cm of add.center, anchor=center] {};
		\node (out) [below=\dstC of add.south, anchor=north] {};
		\node (stretch) [below=2.25cm of in.south, anchor=north] {};

		\draw [->] (in) -- (box1);
		\draw [->] (box1) -- (box2);
		\draw (resid) -- (residr);
		\draw (residr) -- (residbr);
		\draw [->] (residbr) -- (add);
		\draw [->] (box3) -- (add);
		\draw [->] (add) -- (out);
	\end{tikzpicture}
	\caption{Synthesis block (SB)}
	\label{fig:transforms-synthesis-block}
\end{subfigure}
\caption{Transform types. Each layer is specified as follows: convolution type (C refers to convolution, C\textsuperscript{T} to transposed convolution), number of filters, filter size and strides.}
\label{fig:transforms}
\end{figure}
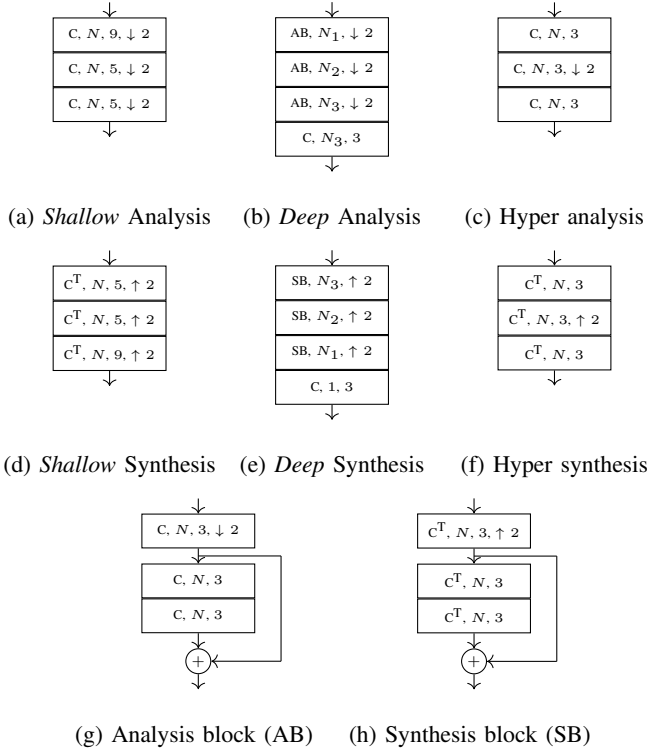

We compare shallow and deep transforms for analysis and synthesis, as illustrated in Fig. \ref{fig:transforms}.
Specifically, we focus on analysis and synthesis transforms and use shallow hyper-analysis and hyper-synthesis transforms (Fig. \ref{fig:transforms-hyper-analysis} and \ref{fig:transforms-hyper-synthesis}).

The transforms based on 3D convolutions and 3D transpose convolutions introduced in \cite{quach_learning_2019} are referred to as \vone{} transforms (Fig. \ref{fig:transforms-analysis-v1} and \ref{fig:transforms-synthesis-v1}).
We introduce \textit{deeper} variants of \vone{} transforms (Fig. \ref{fig:transforms-analysis-v2} and \ref{fig:transforms-synthesis-v2}), which we refer to as \vtwo{} transforms.
These transforms are composed of residual \cite{he_deep_2016} blocks (Fig. \ref{fig:transforms-analysis-block} and \ref{fig:transforms-synthesis-block}) which use skip-connections to prevent issues such as exploding or vanishing gradients.
The skip-connections act as ``shortcuts'' in the network allowing gradients to backpropagate through shorter paths.
We also make them \textit{progressive} by increasing the number of filters progressively as $\NI / 4 = \NII / 2 = \NIII$.
The rationale behind this choice is that the number of filters should compensate the downsampling along the spatial dimensions.
In that way, the capacity at a given layer $\Wi \times \He \times \De \times \N$ decreases more slowly which allows the network to compress information more easily.
In our experiments, we set $\NIII = 64$.

\subsection{Changing the balancing weight in the focal loss (\condalphasafe{})} \label{ssec:focalloss}

\begin{table*}
  \centering
  \begin{tabular}{|l|l|*{8}{r|}}
    \cline{3-8}
    \multicolumn{2}{c|}{} & \multicolumn{6}{c|}{Experimental conditions} \\ \hline
    Point cloud & Metric & \condsequential{} & \condoptimal{} & \condalpha{} & \conddeep{} & \condhyper{} & \condbaseline{} \\ \hline
    \multirow{2}{*}{loot} & D1 & $\bm{5.91}$ / $\bm{7.00}$ & $\mathit{5.84}$ / $\mathit{6.89}$ & $4.05$ / $5.06$ & $2.03$ / $3.67$ & $-0.27$ / $2.26$ & $-0.72$ / $1.88$\\ \cline{2-8}
    & D2 & $\mathit{6.85}$ / $\bm{6.12}$ & $\bm{6.90}$ / $\mathit{6.11}$ & $4.10$ / $3.33$ & $1.44$ / $1.23$ & $-1.81$ / $-0.81$ & $-2.60$ / $-1.40$\\ \hline
    \multirow{2}{*}{redandblack} & D1 & $\bm{5.02}$ / $\bm{6.50}$ & $\mathit{4.81}$ / $\mathit{6.30}$ & $3.28$ / $4.71$ & $1.43$ / $3.45$ & $-0.19$ / $2.58$ & $-0.59$ / $2.01$\\ \cline{2-8}
    & D2 & $\bm{5.93}$ / $\bm{5.65}$ & $\mathit{5.74}$ / $\mathit{5.46}$ & $2.91$ / $2.62$ & $0.55$ / $0.79$ & $-1.73$ / $-0.46$ & $-2.42$ / $-1.18$\\ \hline
    \multirow{2}{*}{longdress} & D1 & $\bm{5.54}$ / $\bm{6.94}$ & $\mathit{5.41}$ / $\mathit{6.79}$ & $3.75$ / $5.10$ & $1.81$ / $3.82$ & $-0.26$ / $2.64$ & $-0.79$ / $2.10$\\ \cline{2-8}
    & D2 & $\bm{6.59}$ / $\bm{6.01}$ & $\mathit{6.52}$ / $\mathit{5.91}$ & $3.90$ / $3.30$ & $1.41$ / $1.36$ & $-1.37$ / $-0.34$ & $-2.20$ / $-1.09$\\ \hline
    \multirow{2}{*}{soldier} & D1 & $\bm{5.55}$ / $\bm{6.91}$ & $\mathit{5.49}$ / $\mathit{6.88}$ & $3.76$ / $5.11$ & $1.88$ / $3.88$ & $-0.28$ / $2.60$ & $-0.77$ / $2.13$\\ \cline{2-8}
    & D2 & $\bm{6.54}$ / $\bm{6.02}$ & $\mathit{6.52}$ / $\bm{6.02}$ & $3.86$ / $\mathit{3.36}$ & $1.39$ / $1.45$ & $-1.54$ / $-0.40$ & $-2.31$ / $-1.06$\\ \hline
    \hline
    \multirow{2}{*}{Average} & D1 & $\bm{5.50}$ / $\bm{6.84}$ & $\mathit{5.39}$ / $\mathit{6.71}$ & $3.71$ / $5.00$ & $1.79$ / $3.71$ & $-0.25$ / $2.52$ & $-0.72$ / $2.03$\\ \cline{2-8}
    & D2 & $\bm{6.48}$ / $\bm{5.95}$ & $\mathit{6.42}$ / $\mathit{5.87}$ & $3.69$ / $3.15$ & $1.20$ / $1.21$ & $-1.61$ / $-0.50$ & $-2.38$ / $-1.18$\\ \hline
  \end{tabular}
  \caption{RD performance for each experimental condition. We specify BD-PSNR values (dB) compared to G-PCC trisoup and G-PCC octree in each cell (trisoup BD-PSNR / octree BD-PSNR). The best values for trisoup and octree are indicated in \textbf{bold} and the second best in \textit{italic}. \condsequential{} consistently outperforms all other conditions.}
  \label{tbl:expmain}
\end{table*}

\begin{table}
  \centering
  \begin{tabular}{|l|l|*{4}{r|}}
    \cline{3-6}
    \multicolumn{2}{c|}{} & \multicolumn{4}{c|}{$\alpha$} \\ \hline
    Point cloud & Metric & $0.90$ & $0.75$ & $0.50$ & $0.25$\\ \hline
    \multirow{2}{*}{loot} & D1 & $2.03$ & $\mathit{4.05}$ & $\bm{4.41}$ & $1.24$\\ \cline{2-6}
    & D2 & $1.44$ & $4.10$ & $\bm{6.47}$ & $\mathit{4.19}$\\ \hline
    \multirow{2}{*}{redandblack} & D1 & $1.43$ & $\bm{3.28}$ & $\mathit{2.24}$ & $-3.70$\\ \cline{2-6}
    & D2 & $0.55$ & $\mathit{2.91}$ & $\bm{5.19}$ & $1.63$\\ \hline
    \multirow{2}{*}{longdress} & D1 & $1.81$ & $\bm{3.75}$ & $\mathit{3.66}$ & $-0.11$\\ \cline{2-6}
    & D2 & $1.41$ & $\mathit{3.90}$ & $\bm{6.28}$ & $3.88$\\ \hline
    \multirow{2}{*}{soldier} & D1 & $1.88$ & $\mathit{3.76}$ & $\bm{4.48}$ & $1.68$\\ \cline{2-6}
    & D2 & $1.39$ & $3.86$ & $\bm{6.32}$ & $\mathit{4.45}$\\ \hline
    \hline
    \multirow{2}{*}{Average} & D1 & $1.79$ & $\bm{3.71}$ & $\mathit{3.70}$ & $-0.22$\\ \cline{2-6}
    & D2 & $1.20$ & $\mathit{3.69}$ & $\bm{6.07}$ & $3.54$\\ \hline
  \end{tabular}
  \caption{Impact of the focal loss $\alpha$ parameter on RD performance. We specify BD-PSNR values (dB) compared to G-PCC trisoup for different $\alpha$ values. The best values are indicated in \textbf{bold} and the second best in \textit{italic}. $\alpha = 0.75$ outperforms all other $\alpha$ values.}
  \label{tbl:expalpha}
\end{table}

When considering point clouds as voxel grids, we observe that most of the space is empty (usually $> 95\%$).
This large class imbalance between occupied voxels and unoccupied voxels is a barrier to effective training.
Indeed, without any countermeasures, the network would converge towards empty outputs only.
In order to resolve this class imbalance issue, we adopt the focal loss \cite{lin_focal_2017} as our distortion loss.

The focal loss is well suited for point clouds since it addresses the class imbalance issue with $\alpha$-balancing.
Moreover, the focal loss differentiates between easy and hard examples using the $\gamma$ parameter.
Specifically, the higher $\gamma$ is, the more hard examples are emphasized.
With $\gamma = 0$, the focal loss becomes equivalent to the weighted binary cross-entropy.

For conciseness, we adopt the following notation.
If $\x = 1$, then $\xt = \x$, $\alphat = \alpha$ and $\xtildet = \xtilde$; otherwise $\xt = 1 - \x$, $\alphat = 1 - \alpha$ and $\xtildet = 1 - \xtilde$.
We then define the focal loss as
\begin{equation}
  \FL(\x, \xtilde) = \alphat \xt (1 - \xtildet)^\gamma \log(\xtildet).
\end{equation}

We study the impact of the focal loss $\alpha$ parameter on RD performance.
The $\alpha$ parameter governs the attention given to occupied voxels and empty voxels.
A high $\alpha$ value makes marking occupied voxels as empty more costly than marking empty voxels as occupied and results in denser reconstructions.
Originally, we picked the same $\alpha$ value ($0.90$) as in \cite{quach_learning_2019}.
This was motivated by the fact that point clouds are often comprised of more than $95\%$ of empty space.

However, we found experimentally that lower $\alpha$ values can actually provide better coding gains.
We hypothesize that this is due to the fact that the default $\gamma = 2$ in the focal loss emphasizes hard examples (occupied voxels) more than easy examples (empty voxels).
Thus, $\gamma = 2$ already alleviates the class imbalance issue to some extent which explains this phenomenon.

\subsection{Optimal thresholding for decoding (\condoptimalsafe{})} \label{ssec:optimal}

For each block, after decoding $y$ (and $z$ for the hyperprior model) into $\xtilde$, we need to convert $\xtilde$ into binary values in order to obtain the decompressed point cloud.
The baseline method (\condbaseline{}) employs a fixed threshold $\thr = 0.5$.
In contrast, we perform this conversion by finding optimal thresholds for each block of voxels. This threshold is transmitted as side information in the bitstream with a small overhead in terms of bitrate.

We formulate optimal thresholding as the problem of finding an optimal threshold $\thrstar$ such that
\begin{equation} \label{eq:optthr}
  \thrstar = \argmin_\thr \dist(\x, \Hea(\xtilde - \thr))
\end{equation}
where $\dist$ is a distortion metric and $\Hea(\x)$ is the heaviside step function (equal to $1$ when $\x \geq 0$ and $0$ otherwise).

\subsection{Sequential training (\condsequentialsafe{})} \label{ssec:sequential}

We train compression models for each RD tradeoff using a corresponding $\lambda$ value.
This allows for transforms and entropy models to be specialized for this particular tradeoff resulting in better RD performance.
Unfortunately, using this independent training scheme, we need to train one model for each tradeoff.

To alleviate this issue, we propose a novel sequential training scheme that speeds up training significantly and improves RD performance.
The core idea of this scheme is to use previously trained neural network weights as a starting point for new neural networks.
Essentially, given a set of $\lambda$ tradeoffs, we first train $\lambda_1$.
Then, for each subsequent model, we train $\lambda_i$ using the trained weights of $\lambda_{i-1}$.

In this training scheme, we proceed to train the different tradeoffs in descending order.
That is, we first train a low distortion, high bitrate model.
Then, for each subsequent model, we progressively lower the bitrate while trying to minimize the increase in distortion.

\section{Experiments}

\begin{figure*}
  \centering
  \includegraphics[width=0.9\textwidth]{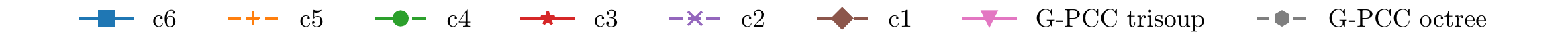}
  \begin{subfigure}{0.24\textwidth}
    \includegraphics[width=\linewidth]{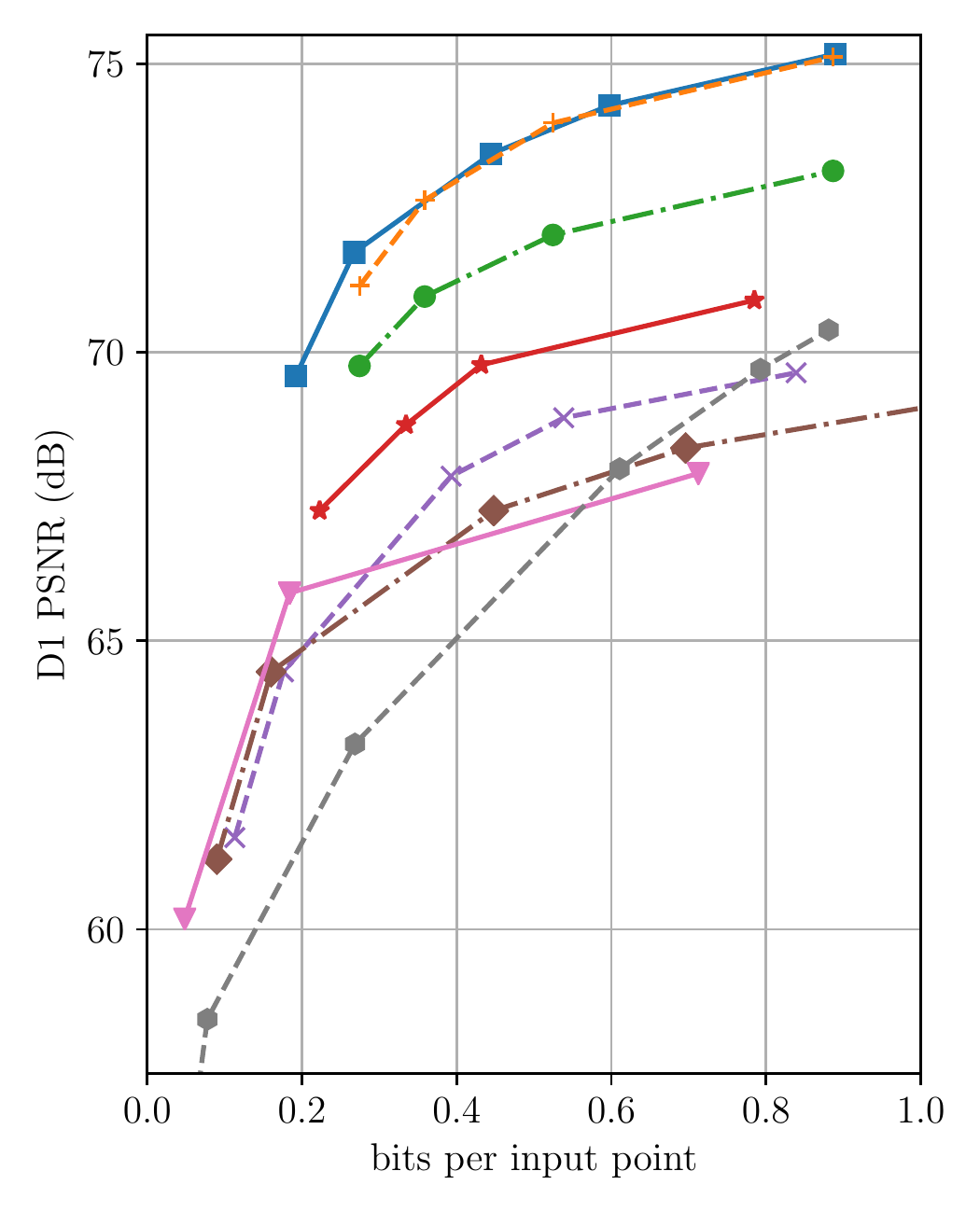}
    \includegraphics[width=\linewidth]{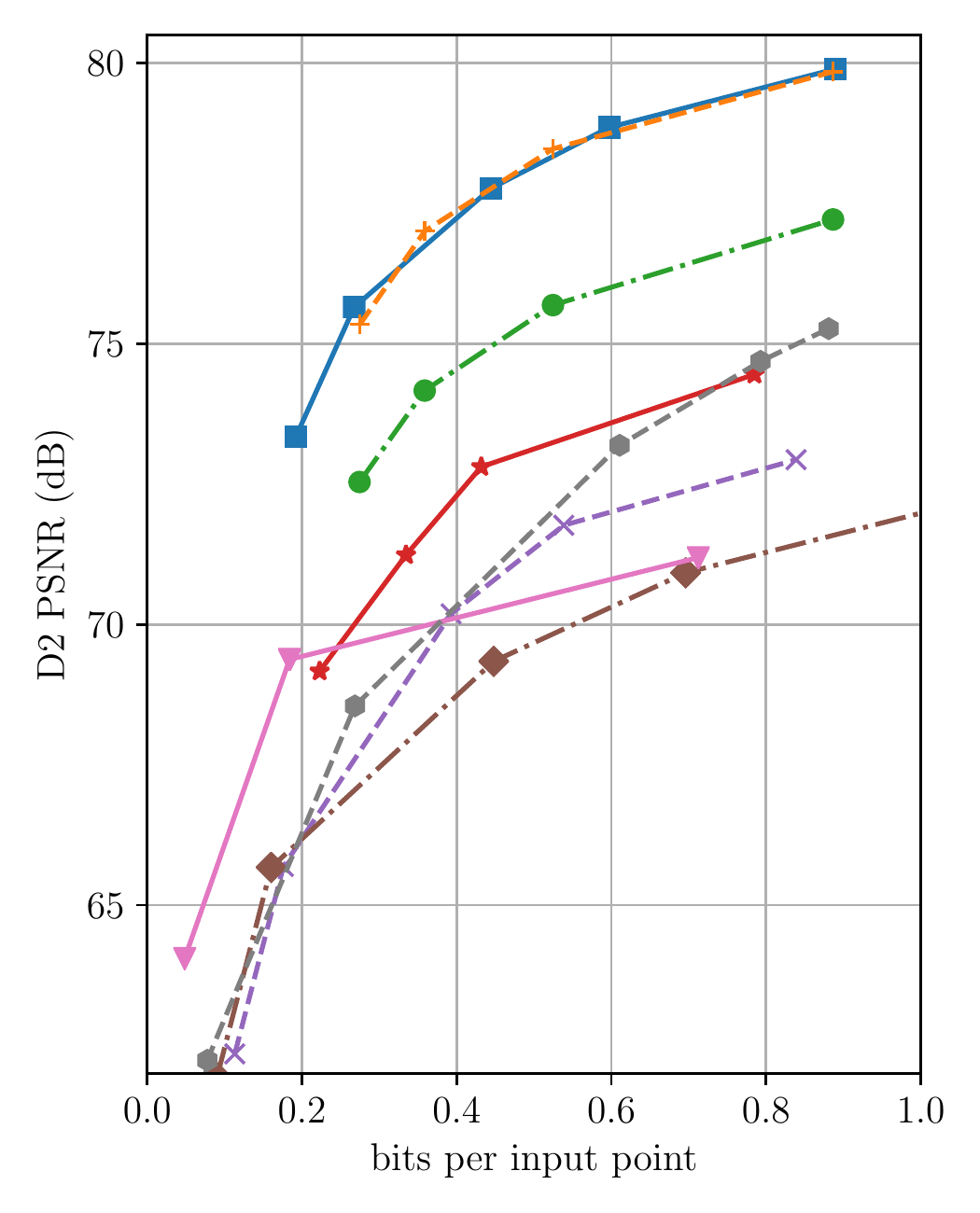}
    \caption{``loot\_vox10\_1200''}
  \end{subfigure}
  \begin{subfigure}{0.24\textwidth}
    \includegraphics[width=\linewidth]{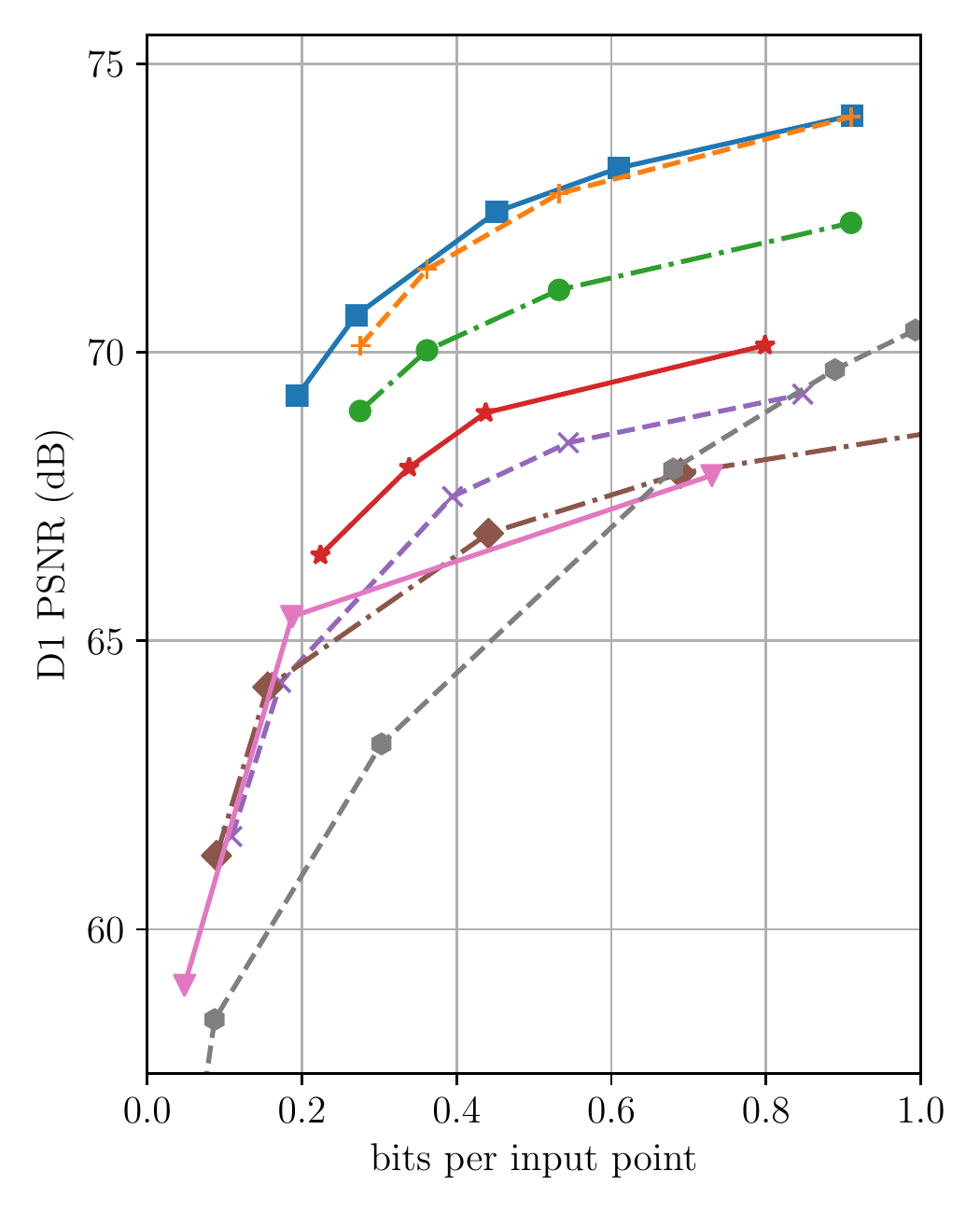}
    \includegraphics[width=\linewidth]{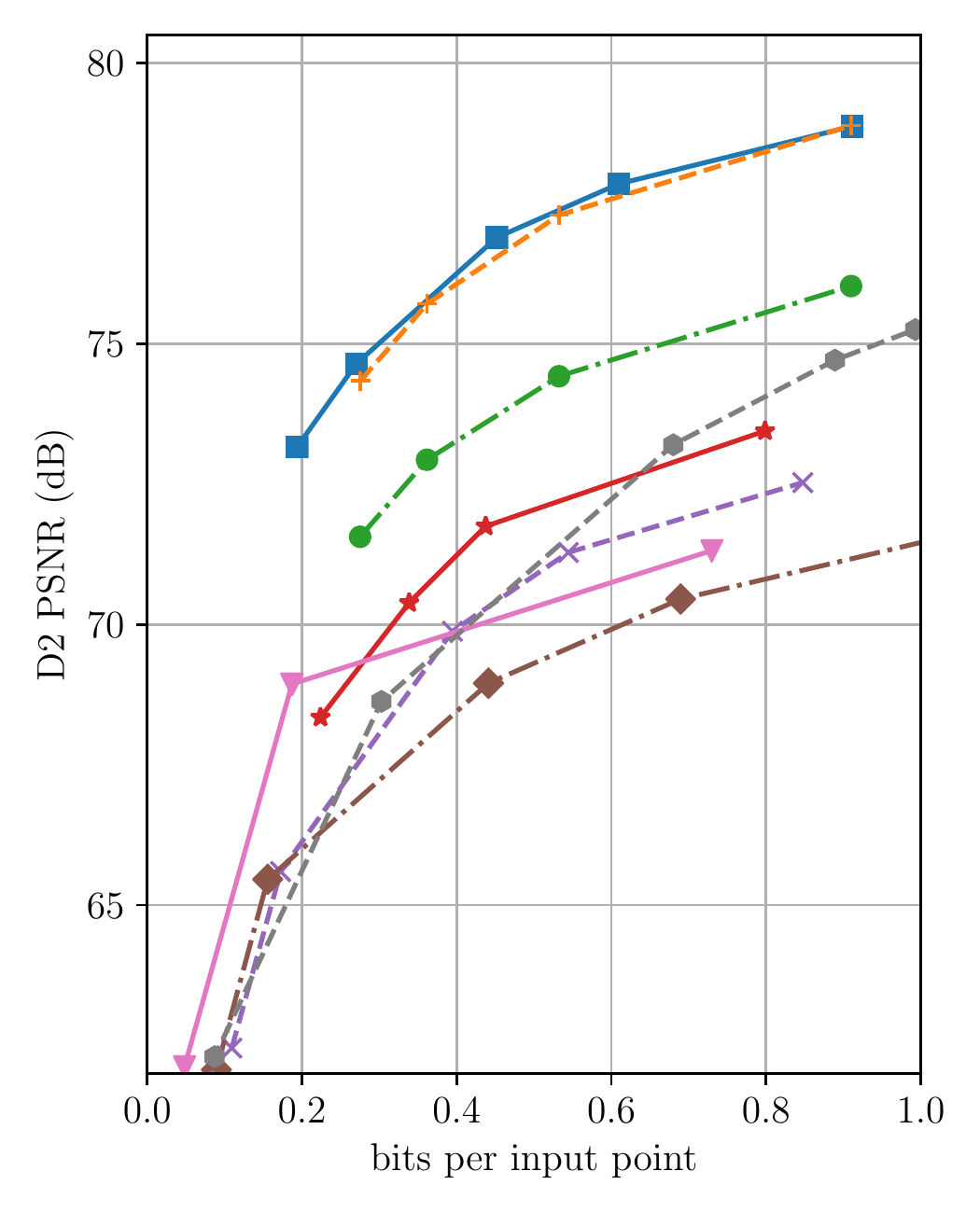}
    \caption{``redandblack\_vox10\_1490''}
  \end{subfigure}
  \begin{subfigure}{0.24\textwidth}
    \includegraphics[width=\linewidth]{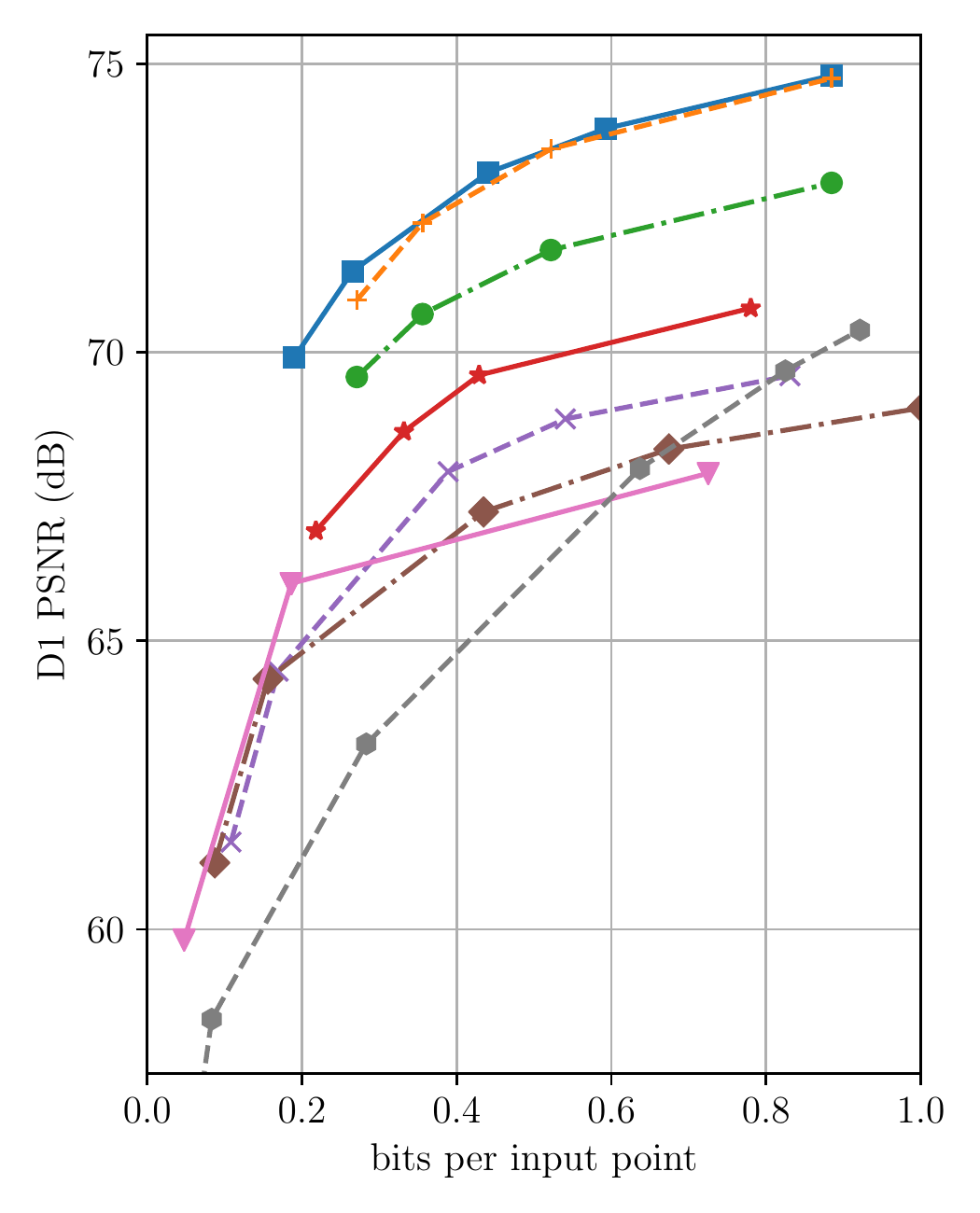}
    \includegraphics[width=\linewidth]{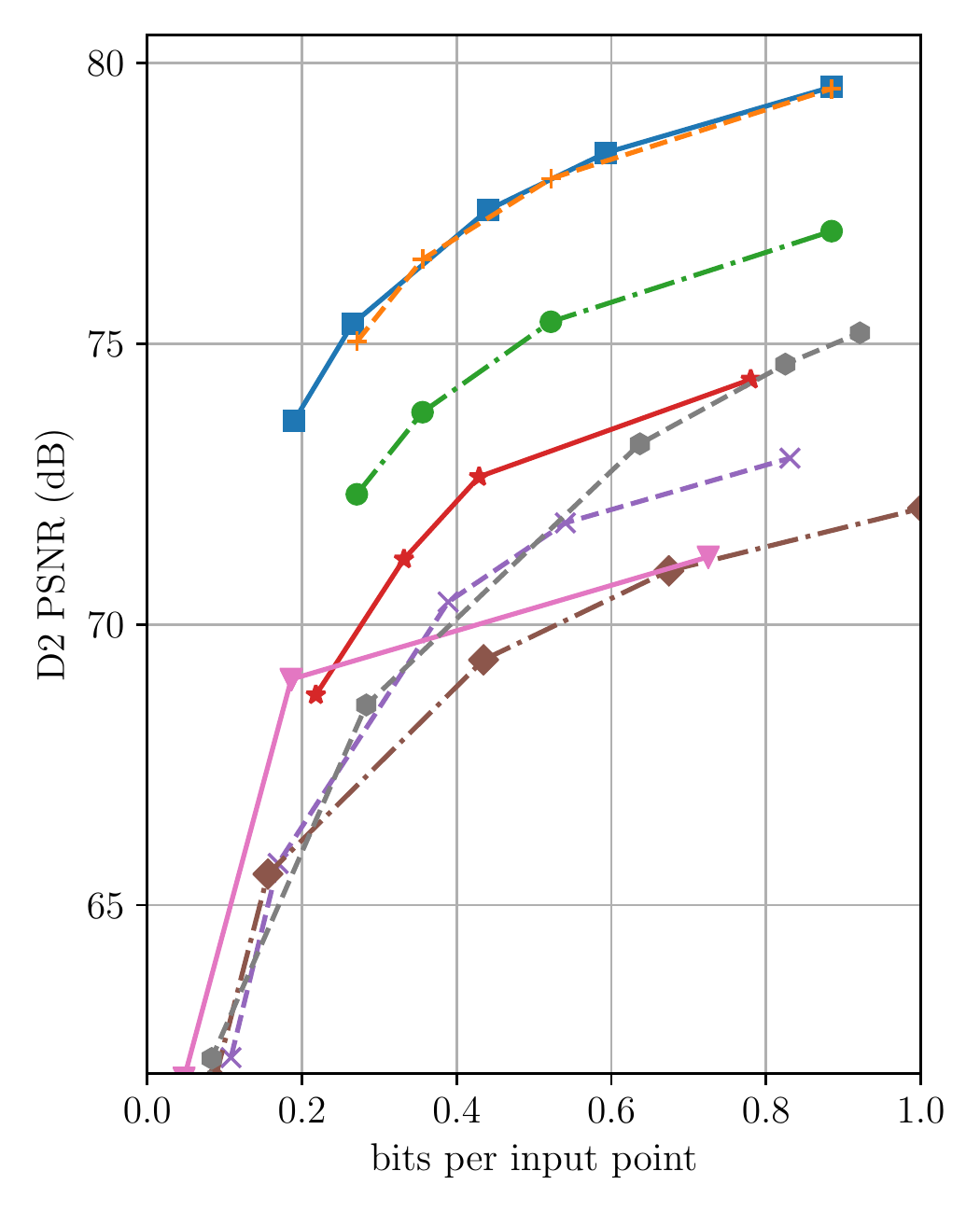}
    \caption{``longdress\_vox10\_1300''}
  \end{subfigure}
  \begin{subfigure}{0.24\textwidth}
    \includegraphics[width=\linewidth]{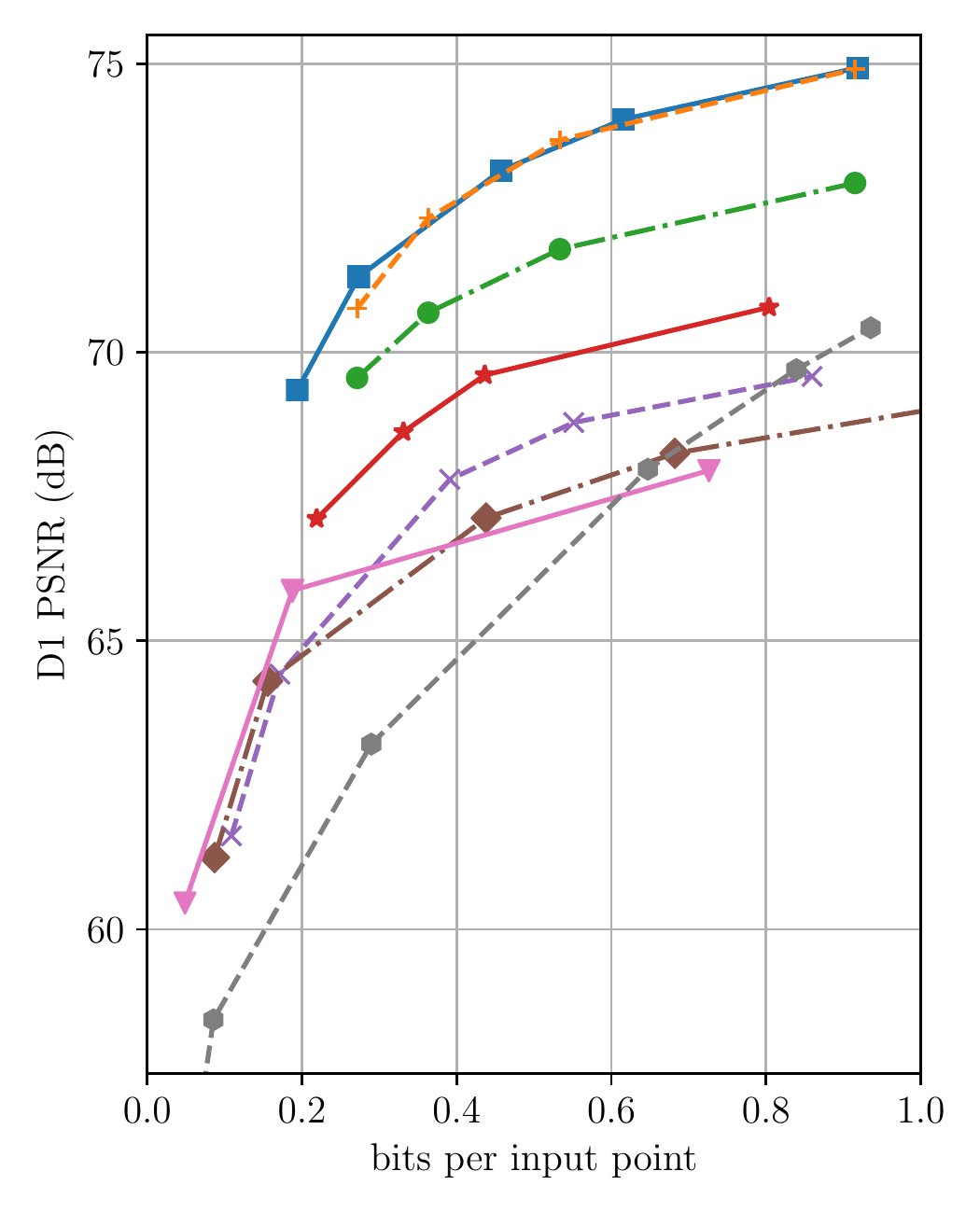}
    \includegraphics[width=\linewidth]{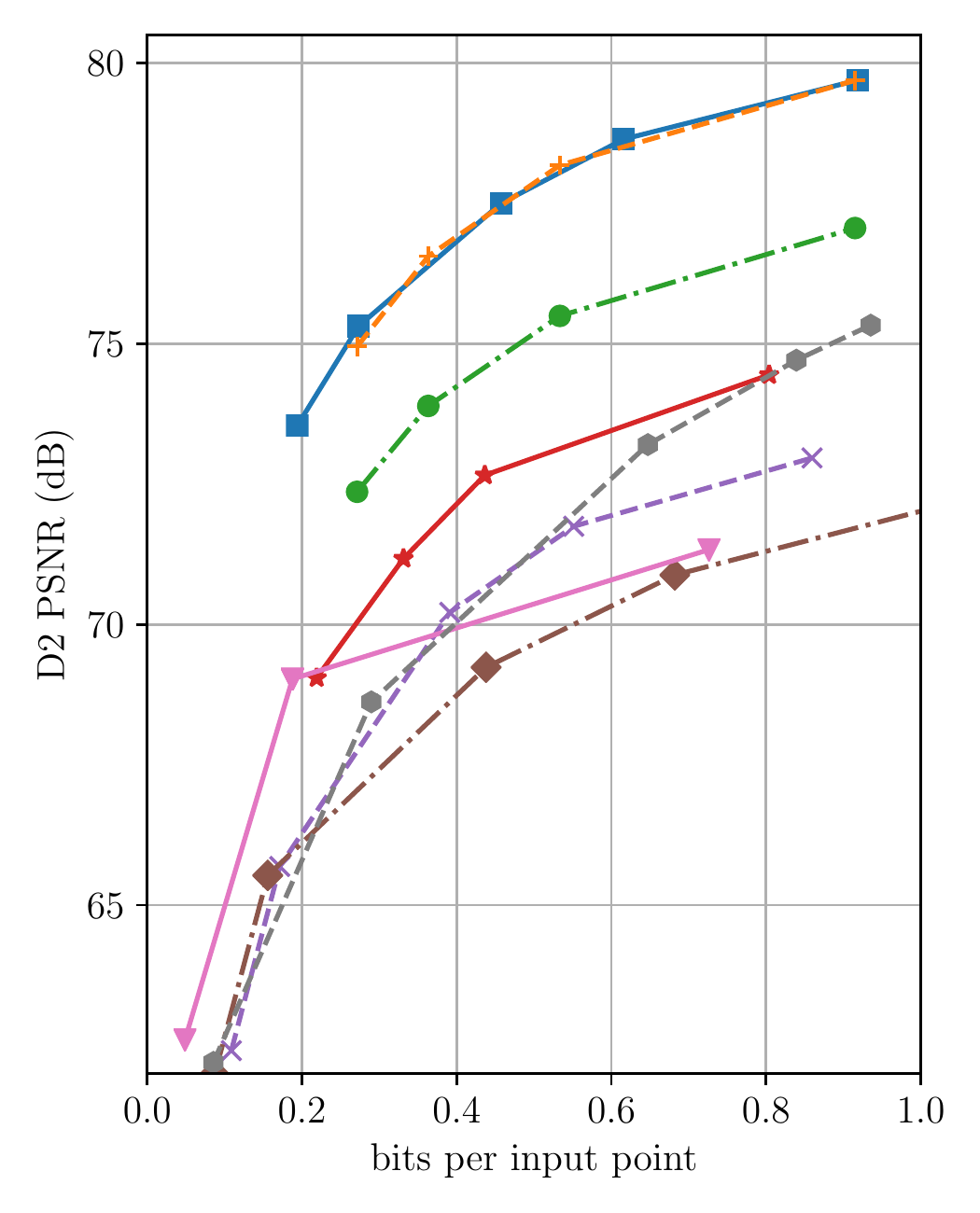}
    \caption{``soldier\_vox10\_0690''}
  \end{subfigure}
  \caption{RD curves for each condition in Table \ref{tbl:expmain}. \condsequential{} consistently outperforms G-PCC trisoup and G-PCC octree.}
  \label{fig:rdcurvesmain}
\end{figure*}

\begin{figure*}
  \centering
  \begin{subfigure}{0.18\textwidth}
    \includegraphics[width=0.98\linewidth]{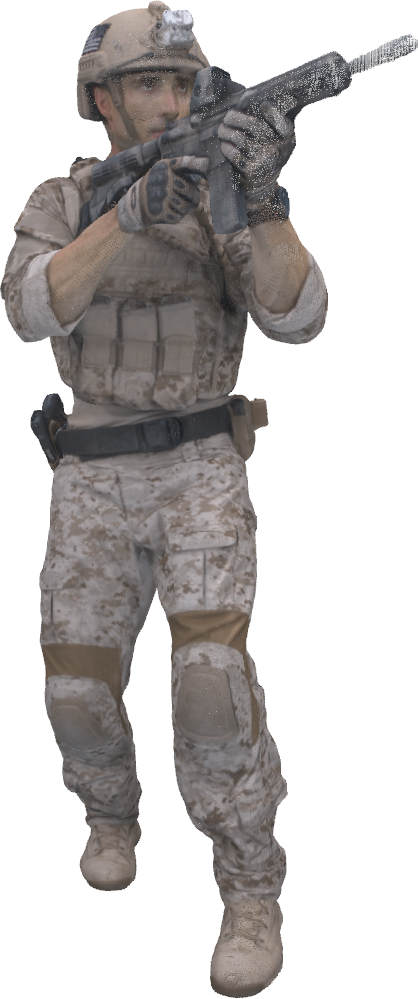}
    \caption{Original}
  \end{subfigure}
  \begin{subfigure}{0.36\textwidth}
    \includegraphics[width=0.49\linewidth]{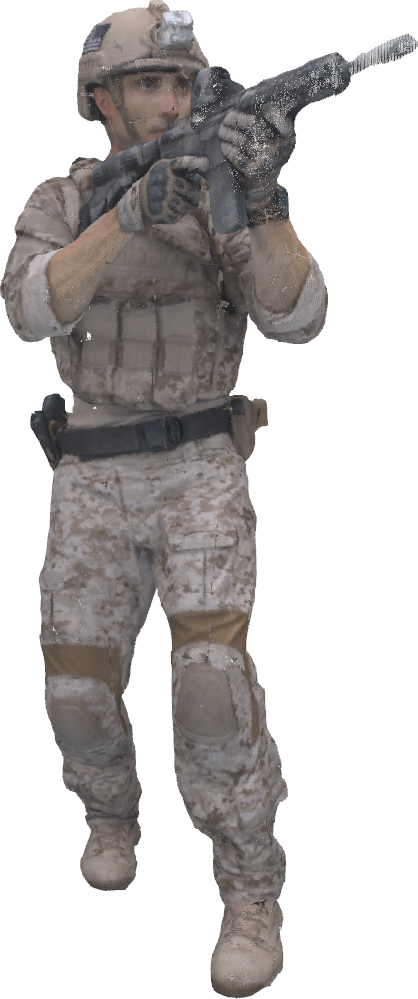}
    \includegraphics[width=0.49\linewidth]{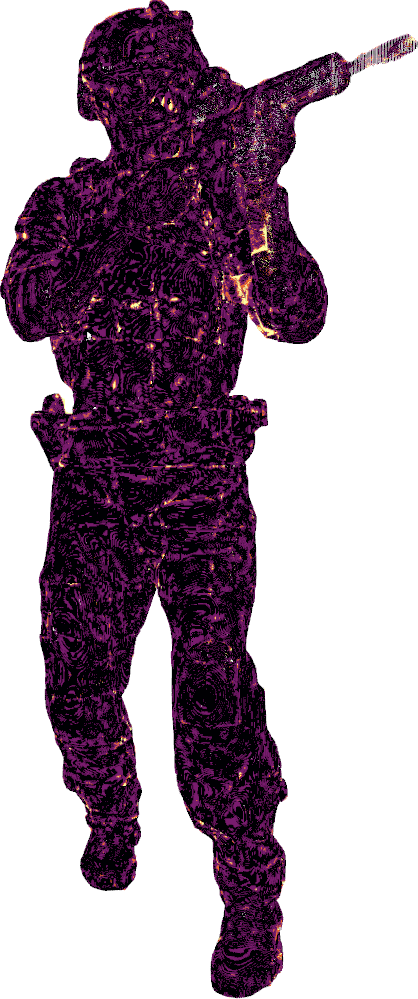}
    \caption{\condsequential{} (D1 $69.59$dB, $0.194$ bpp)}
  \end{subfigure}
  \begin{subfigure}{0.36\textwidth}
    \includegraphics[width=0.49\linewidth]{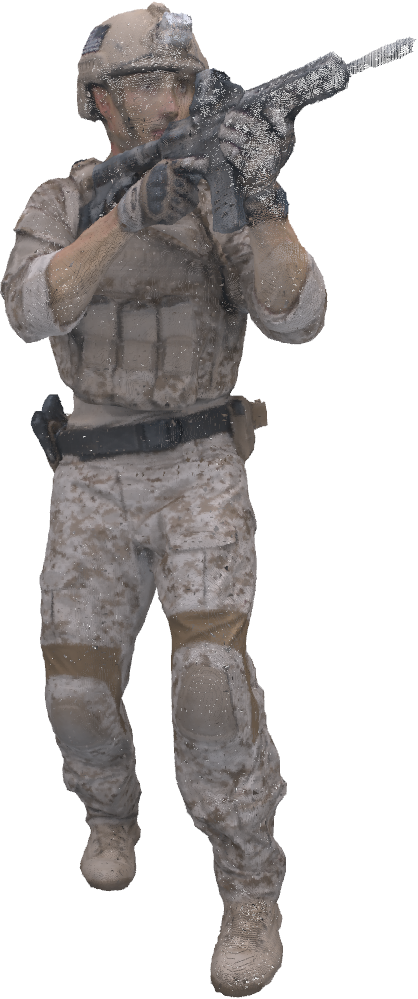}
    \includegraphics[width=0.49\linewidth]{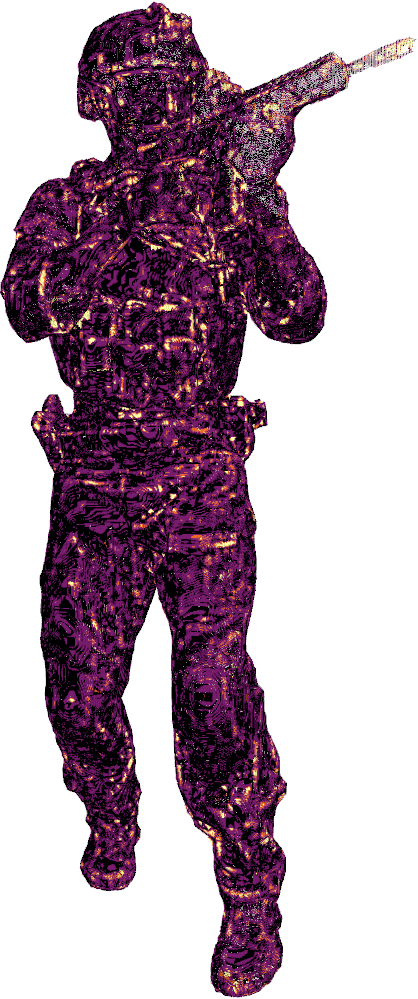}
    \caption{G-PCC Trisoup (D1 $65.87$dB, $0.188$ bpp)}
  \end{subfigure}
  \begin{subfigure}{0.07\textwidth}
    \includegraphics[width=\linewidth]{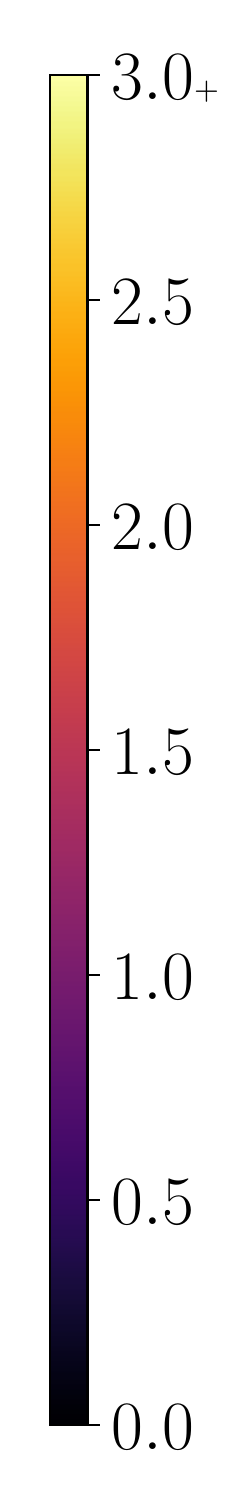}
    \vspace{0.3cm}
  \end{subfigure}
  \caption{Qualitative evaluation on ``soldier\_vox10\_0690''. For \condsequential{} and G-PCC Trisoup, we show the decompressed point cloud and its D1 squared errors. The errors are displayed according to the color scale on the right and are truncated to the 99th percentile ($3.0$). In parentheses, we specify the D1 PSNR along with the number of bits per input point (bpp).}

\end{figure*}

In this paper, we evaluate the six different improvement strategies described in Section~\ref{sec:proposed} and summarized in Table \ref{tbl:expcond}. The BD-PSNR gains are reported in Table \ref{tbl:expmain}.

\subsection{Experimental setup}

We perform our experiments on a subset of four point clouds specified in the MPEG CTCs \cite{noauthor_common_2020}.
Namely, ``longdress\_vox10\_1300'', ``loot\_vox10\_1200'', ``redandblack\_vox10\_1490'', ``soldier\_vox10\_0690''.
We also refer to these four point clouds as ``longdress'', ``loot'', ``redandblack'' and ``soldier''.

We train our models on a subset of the ModelNet40 dataset.
First, we sample the dataset into voxelized point clouds with resolution $512$ and select the $200$ largest point clouds.
Then, we divide these point clouds into blocks with resolution $64$ and select the $4000$ largest blocks.
This produces a small dataset containing rich point clouds, accelerates dataset loading time and reduces memory footprint when training.
We perform training with $\lambda$ values ranging from $5\times 10^{-6}$ to $3\times 10^{-4}$.

We evaluate the different conditions using G-PCC trisoup and octree as baselines.
Specifically, we use G-PCC v10.0 (released in May 2020) with the included configurations, ``mpeg-pcc-dmetric'' v0.12.3 for D1 and D2 metrics, Python 3.6.9 and TensorFlow 1.15.0 with the Adam optimizer \cite{kingma_adam:_2014}.

\subsection{Experimental results}

In Fig. \ref{fig:rdcurvesmain} and Table \ref{tbl:expmain}, we observe that each condition is a net improvement over previous ones.
\condsequential{} outperforms G-PCC trisoup with an average BD-PSNR of $5.50$ dB on D1 and $6.48$ dB on D2 and outperforms G-PCC octree with an average BD-PSNR of $6.84$ dB on D1 and $5.95$ dB on D2.
Note that the lowest bitrate point for \condsequential{} is not included in BD-PSNR computations in order to keep integration intervals consistent and keep BD-PSNRs comparable across different conditions.

We also observe that \condoptimal{} (optimal thresholding) is especially beneficial for the point-to-plane metric (D2) with an improvement of $1.68$ dB for D1 and $2.73$ dB for D2 compared to \condalpha{}.
Indeed, optimal thresholding provides optimal sets of thresholds for D1 and D2 yielding two separate reconstructions.

\subsection{Ablation study}

In this subsection, we present BD-PSNR values when compared to G-PCC trisoup.
The hyperprior model (\condhyper{}) results in an improvement of $0.47$ dB for D1 and $0.77$ dB for D2 compared to \condbaseline{}.
Adding \vtwo{} transforms (\conddeep{}) further improves D1 by $2.04$ dB and D2 by $2.81$ dB compared to \condhyper{}.

In Table \ref{tbl:expalpha}, we observe that setting $\alpha = 0.75$ for D1 and $\alpha = 0.50$ for D2 increases RD performance significantly for all point clouds.
The average BD-PSNR for $\alpha = 0.75$ is $3.71$ dB for D1 and $3.69$ dB for D2.
Also, the average BD-PSNR for $\alpha = 0.50$ is $3.70$ dB for D1 and $6.07$ dB for D2.
Indeed, higher $\alpha$ values lead to denser reconstructions which are favored by D1 and lower $\alpha$ values to sparser ones which are favored by D2.
We select $\alpha = 0.75$ (\condalpha{}) as we have found experimentally that it performs better when associated with optimal thresholding.
Compared to $\alpha = 0.90$ (\conddeep{}), $\alpha = 0.75$ brings an improvement of $1.92$ dB for D1 and $2.49$ dB for D2 .

Then, we use optimal thresholding (\condoptimal{}) with the point-to-point (D1) and point-to-plane (D2) objective metrics.
As a result, we obtain two point clouds respectively optimized with D1 and D2.
Also, we encode thresholds on $8$ bits with $256$ uniformly distributed threshold values between $0$ and $1$.
Optimal thresholding (\condoptimal{}) results in an improvement of $1.68$ dB for D1 and $2.73$ dB for D2 compared to \condalpha{}.

\begin{figure}
    \centering
    \begin{subfigure}{\columnwidth}
        \centering
        \includegraphics[width=0.48\linewidth]{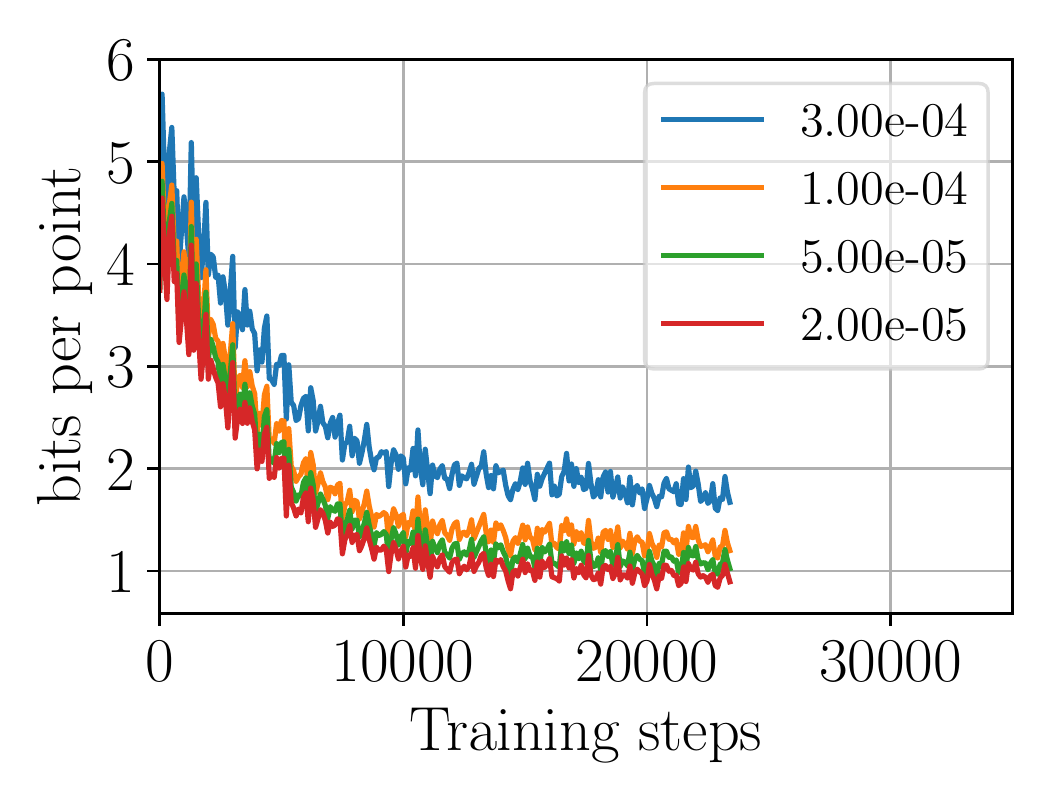}
        \includegraphics[width=0.48\linewidth]{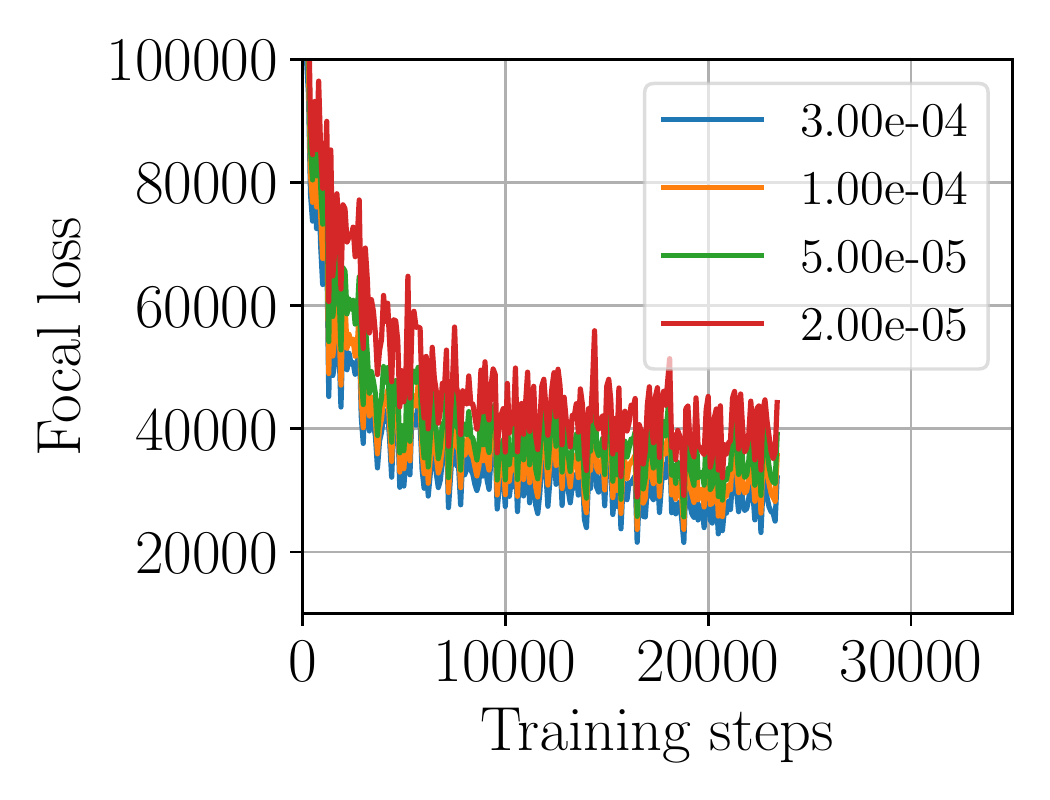}
        \caption{Independent training.}
    \end{subfigure}
    \begin{subfigure}{\columnwidth}
        \centering
        \includegraphics[width=0.48\linewidth]{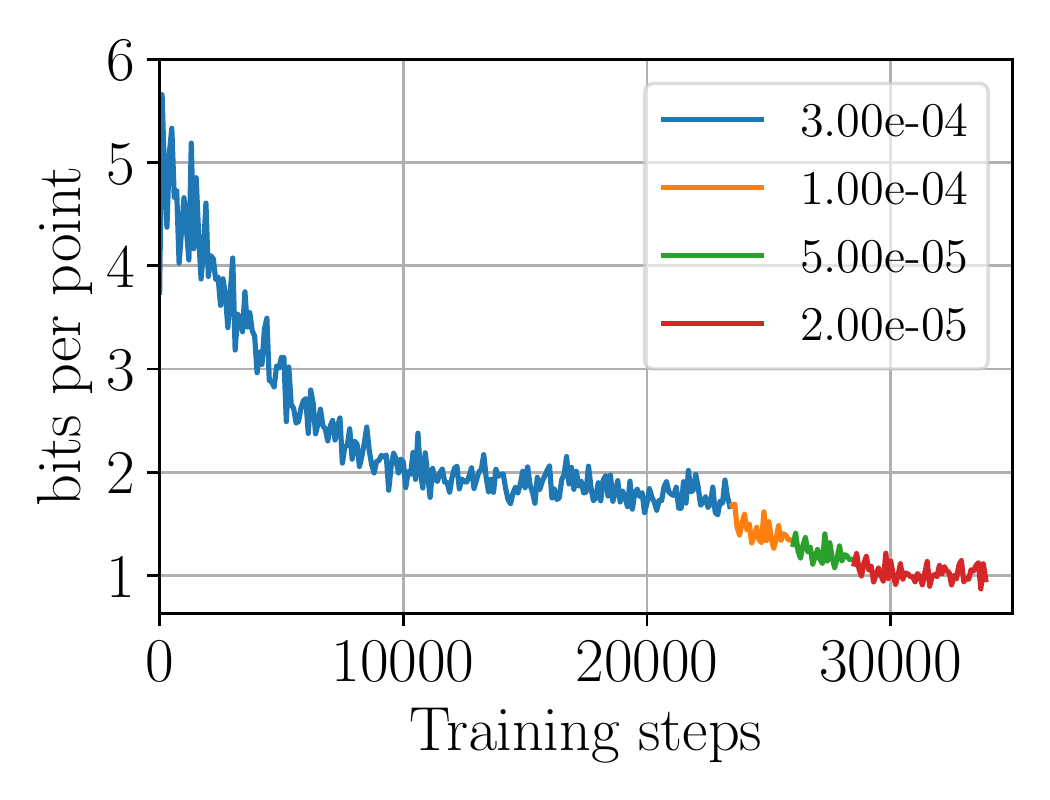}
        \includegraphics[width=0.48\linewidth]{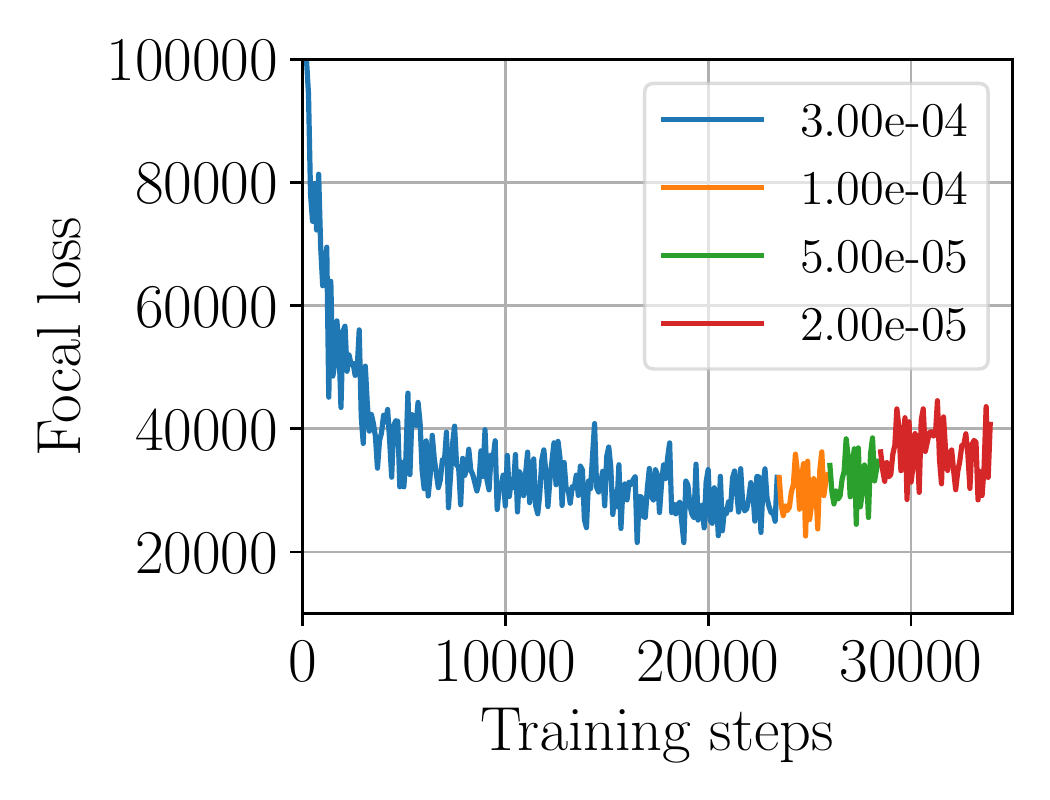}
        \caption{Sequential training.}
    \end{subfigure}
    \caption{Bits per point and focal loss when training independently and sequentially. Sequential training is more efficient as it reuses previously trained models to train subsequent ones.}
    \label{fig:sequential}
\end{figure}

Training DPCC models is time consuming as shown in Fig. \ref{fig:sequential}.
Indeed, the \condoptimal{} condition requires 4 hours of training resulting in a total of 16 hours for four models on an Nvidia GeForce GTX 1080 Ti.
With sequential training (\condsequential{}), these models train in 30 to 60 minutes instead of 4 hours which is up to $8$ times faster.
In addition, this results in an improvement of $0.11$ dB for D1 and $0.06$ dB for D2 compared to \condoptimal{}.

\section{Conclusion}

We propose a set of key performance factors for DPCC and we present an extensive ablation study on the individual impact of these factors.
More precisely, we provide insights on the individual impact of scale hyperprior models, \vtwo{} transforms, the focal loss $\alpha$ value, optimal thresholding and sequential training.
We analyze each of these factors in order to provide a better understanding about why they improve RD performance. 
The final model (\condsequential{}) outperforms G-PCC trisoup with an average BD-PSNR of $5.50$ dB on D1 and $6.48$ dB on D2 and outperforms G-PCC octree with an average BD-PSNR of $6.84$ dB on D1 and $5.95$ dB on D2.

\bibliographystyle{IEEEtran}
\bibliography{main}

\end{document}